\documentclass[conference]{IEEEtran}
\IEEEoverridecommandlockouts

\usepackage{cite}
\usepackage{braket}
\usepackage{eso-pic}
\usepackage[hidelinks]{hyperref}
\usepackage{amsmath,amssymb,amsfonts}
\usepackage{algorithmic}
\usepackage{graphicx}
\usepackage{textcomp}
\usepackage{xcolor}
\def\BibTeX{{\rm B\kern-.05em{\sc i\kern-.025em b}\kern-.08em
    T\kern-.1667em\lower.7ex\hbox{E}\kern-.125emX}}

\begin{document}

\title{A Novel Approach to Explainable AI with Quantized Active Ingredients in Decision Making}

\author{\IEEEauthorblockN{A.M.A.S.D. Alagiyawanna}
\IEEEauthorblockA{\textit{Department of Computational Mathematics} \\
\textit{University of Moratuwa}\\
Sri Lanka \\
alagiyawannaamasd.21@uom.lk}\\

\IEEEauthorblockN{Thushari Silva}
\IEEEauthorblockA{\textit{Department of Computational Mathematics} \\
\textit{University of Moratuwa}\\
Sri Lanka \\
thusharip@uom.lk}

\and

\IEEEauthorblockN{Asoka Karunananda}
\IEEEauthorblockA{\textit{Department of Computational Mathematics} \\
\textit{University of Moratuwa}\\
Sri Lanka \\
asokakaru@uom.lk}\\

\IEEEauthorblockN{A. Mahasinghe}
\IEEEauthorblockA{\textit{Department of Mathematics} \\
\textit{University of Colombo}\\
Sri Lanka \\
anuradhamahasinghe@maths.cmb.ac.lk}
}

\maketitle

\begin{center}
\small
This is a preprint of a paper accepted at the 8th SLAAI International Conference on Artificial Intelligence (SLAAI-ICAI 2025) and published by IEEE.\\
The final published version is available at
\href{https://doi.org/10.1109/SLAAI-ICAI68534.2025.11318441}{10.1109/SLAAI-ICAI68534.2025.11318441}.
\end{center}
\vspace{1em}

\begin{abstract}
Artificial Intelligence (AI) systems have shown good success at classifying. However, the lack of explainability is a true and significant challenge, especially in high-stakes domains, such as health and finance, where understanding is paramount. We propose a new solution to this challenge: an explainable AI framework based on our comparative study with Quantum Boltzmann Machines (QBMs) and Classical Boltzmann Machines (CBMs). We leverage principles of quantum computing within classical machine learning to provide substantive transparency around decision-making. The design involves training both models on a binarised and dimensionally reduced MNIST dataset, where Principal Component Analysis (PCA) is applied for preprocessing. For interpretability, we employ gradient-based saliency maps in QBMs and SHAP (SHapley Additive exPlanations) in CBMs to evaluate feature attributions.QBMs deploy hybrid quantum-classical circuits with strongly entangling layers, allowing for richer latent representations, whereas CBMs serve as a classical baseline that utilises contrastive divergence. Along the way, we found that QBMs outperformed CBMs on classification accuracy (83.5\% vs. 54\%) and had more concentrated distributions in feature attributions as quantified by entropy (1.27 vs. 1.39). In other words, QBMs not only produced better predictive performance than CBMs, but they also provided clearer identification of ``active ingredient'' or the most important features behind model predictions. To conclude, our results illustrate that quantum-classical hybrid models can display improvements in both accuracy and interpretability, which leads us toward more trustworthy and explainable AI systems.
\end{abstract}

\begin{IEEEkeywords}
quantum boltzmann machine, classical boltzmann machine, quantum parallelism, explainable artificial intelligence, gradient-based saliency, shap, feature importance, quantum machine learning, principal component analysis, model interpretability, entropy-based explainability
\end{IEEEkeywords}

\section{Introduction}
Machine learning is an area of AI that involves computers learning from data, essentially identifying patterns for prediction or decision-making without undergoing rigorous programming. An ML algorithm can be roughly classified into three categories: supervised learning, where models are trained on labelled data; unsupervised learning, where models try to identify the patterns without any labelling; and reinforcement learning, where the interaction between the agent and environment is marked by trial and error \cite{el2015machine}.

Among the available types of ML models, Boltzmann Machines are considered energy models that learn complex probability distributions on data. A Classical Boltzmann Machine (CBMs) consists of two types of units: a visible layer and a hidden layer, which are symmetrically connected and trained with algorithms \cite{hinton2007boltzmann}. CBMs are, therefore, ideally suited to learning the hidden structure and uncertainty in data; hence, they are meant for unsupervised learning. One major drawback of these models and most deep learning models is the lack of explainability. In high-stakes areas such as healthcare, finance, and law, it is essential to understand why a given model arrives at a decision, not just what that particular decision is.

\begin{figure}[htbp]
\centerline{\fbox{\includegraphics[scale=0.35]{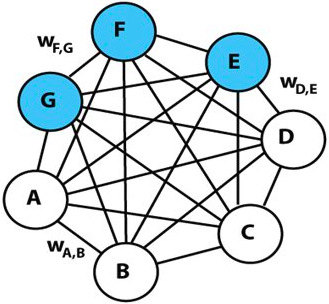}}}
\caption{ A graphical representation of an example Boltzmann machine. In this example, there are 3 hidden units and 4 visible units \cite{patel2020overview}.}
\label{fig1}
\end{figure}

The basis of Quantum Computing unfolds into new possibilities for quantum machine learning phenomena, including superposition, entanglement, and quantum parallelism. Superposition means a qubit can be in several states ($\ket{0}$ and $\ket{1}$) at one time, unlike classical bits, where it can only be 0 or 1. This allows a quantum system to process multiple possibilities simultaneously. Qubits in these states are called ``Ket 0'' and ``Ket 1'' in Dirac notation \cite{alagiyawanna2024enhancing}. In vector notation,

\begin{equation}
    \ket{0} =  
    \begin{bmatrix}
    1 \\
    0 
    \end{bmatrix}
    \ket{1} =  
    \begin{bmatrix}
    0 \\
    1 
    \end{bmatrix}
\end{equation}

\begin{figure}[htbp]
\centerline{\fbox{\includegraphics[scale=0.20]{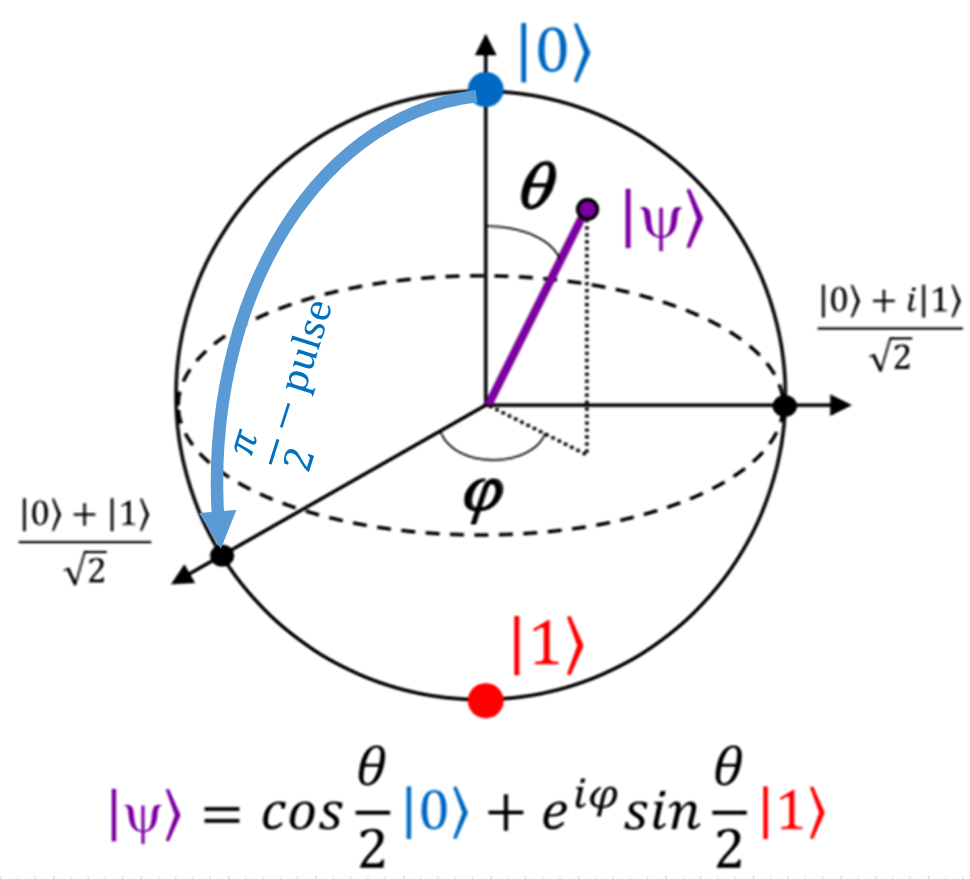}}}
\caption{ Visualization of a qubit using bloch sphere representation \cite{jazaeri2019review}.}
\label{fig2}
\end{figure}

Entanglement is a quantum property in which the state of a qubit is correlated to the state of another-given that they are no matter how far apart the two may be-in which altering one will directly alter the other. These unique properties allow Quantum Boltzmann Machines (QBMs)  \cite{amin2018quantum} to represent richer probability distributions with potentially fewer parameters than classical models. In our work, we integrate quantum computing with classical machine learning by using a hybrid quantum-classical model. The quantum layers are constructed using PennyLane \cite{bergholm2018pennylane}, a framework that simulates quantum circuits and enables automatic differentiation. These circuits contribute to feature learning through gradient-based saliency maps extracted from the quantum latent space, enabling us to interpret the most influential features behind model predictions.

Quantum parallelism builds on superposition by applying a quantum operation to a superposed state, the computation is effectively carried out on all the superposed inputs at once. This allows a quantum system to explore many computational paths in parallel, offering an exponential speed-up for certain problems.

Explainability in classical models is often offered by tools such as SHAP (SHapley Additive exPlanations) \cite{NIPS2017_7062}, which try to assign importance scores to input features by assuming that each feature contributes independently to the output. That independence assumption does not hold for CBMs, in which nodes are tightly interlinked and rely on a lot of joint feature interactions. Hence, the SHAP framework may not be able to provide relevant or meaningful attributions to CBMs. Conversely, SHAP would be even less applicable for QBMs because of the non-classical nature of quantum states and the lack of discrete, interpretable paths through features. In QBMs, qubits are often entangled, meaning the state of one qubit depends on others, violating SHAP’s assumption of feature independence. Therefore, instead of SHAP, we rely on quantum gradients extracted from the latent layer. These gradients naturally reflect how sensitive the quantum model’s output is to each input feature, providing a more compatible and meaningful approach to interpretability in quantum systems.

\begin{figure}[htbp]
\centerline{\fbox{\includegraphics[scale=0.15]{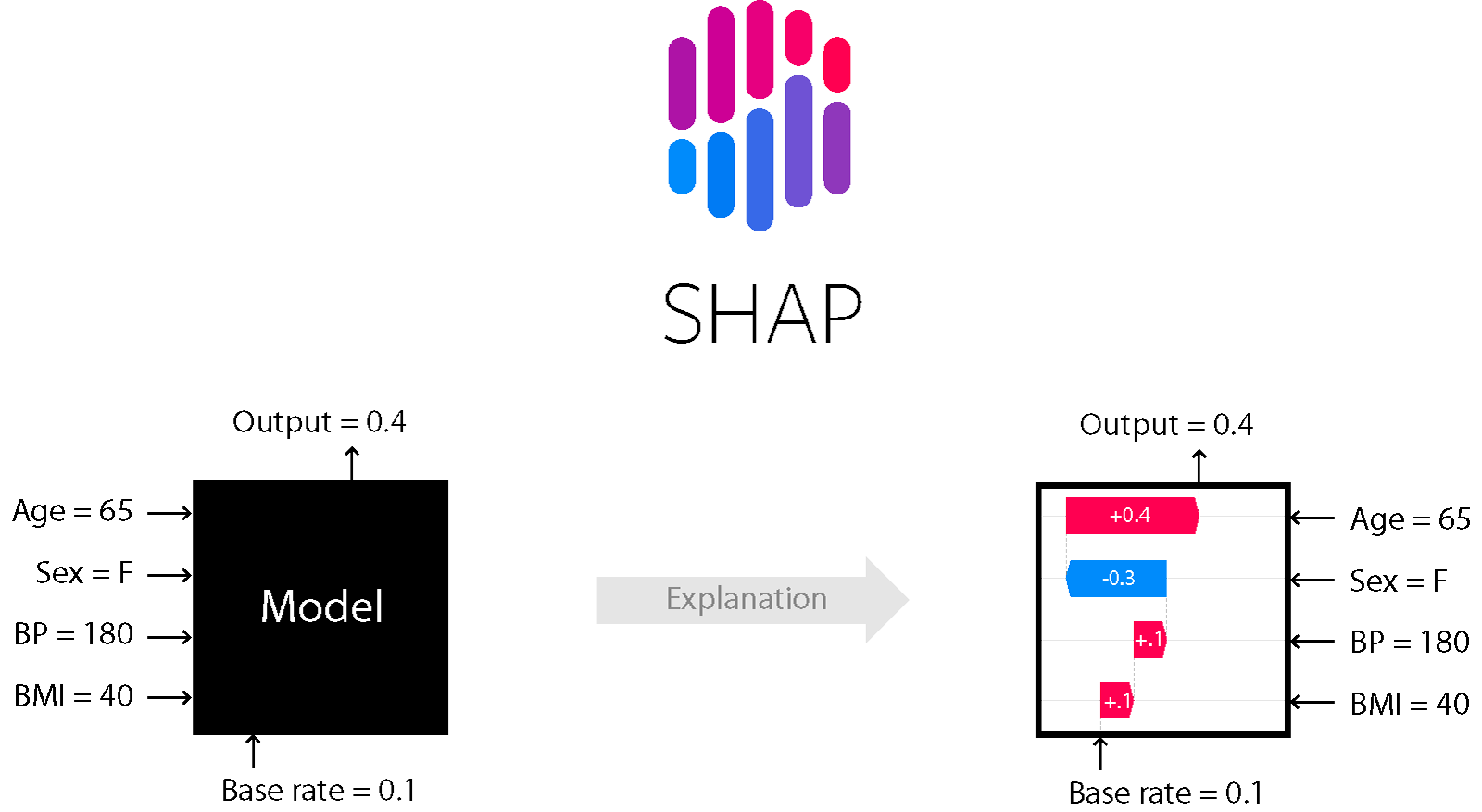}}}
\caption{ Representation of how shap works \cite{NIPS2017_7062}.}
\label{fig4}
\end{figure}

To better understand how models make decisions, we are inspired by the concept of active ingredients that determine the overall characteristics of a natural system.  For instance, in medicine, the active ingredients of a painkiller would be paracetamol, among other components.
In food, it could be curcumin in turmeric, known for its health benefits. Inspired by this, we use the term active ingredient in AI to mean the most important input features that strongly influence the model’s output. In our research, we want to see which features are truly driving the decisions made by the model, just like how doctors want to know which part of a drug is helping the patient. Our goal is to find out whether QBMs can better highlight these key features ``active ingredients'' compared to CBMs. This idea guides our approach to evaluating explainability in both classical and quantum models.

The upcoming sections of the paper are organised as follows. Section II describes the methodology of this research, covering the Approach, reducing dimensionality, Quantum Hidden Layer Design, Design of Classical Boltzmann Machine and Evaluation of QBM. Section III analyses and interprets the results generated by the experiments. Section IV presents the conclusion.

\section{Methodology}

\subsection{Approach}
To explore explainability in machine learning models, our approach integrates quantum computing with classical Boltzmann Machines through a 4-step pipeline: \textit{preparing and reducing input data}, \textit{designing a quantum hidden layer}, \textit{constructing a hybrid Quantum Boltzmann Machine architecture}, and \textit{analysing feature importance using gradient-based saliency maps}.

\begin{itemize}
    \item Preparing and reducing input data: Selecting only two digits (0 and 1) and reducing the input dimensionality. The dimensionality of raw data will be reduced as follows,
    \begin{equation}
        Z = U_K * X
    \end{equation}
    Where Z is the reduced-dimension data, X is the original data, ${U_k}$ is a matrix containing the first k principal components \cite{mackiewicz1993principal}.
    \item Designing a quantum hidden layer: A quantum circuit is built with strongly entangling layers, allowing qubits to capture complex feature interactions through superposition and entanglement.
    \begin{equation}
    \mathbf{z} = \left[ \langle Z_1 \rangle, \langle Z_2 \rangle, \dots, \langle Z_n \rangle \right] = \left\langle 0 \middle| U^\dagger(\mathbf{x}, \boldsymbol{\theta}) Z_i U(\mathbf{x}, \boldsymbol{\theta}) \middle| 0 \right\rangle
    \end{equation} 
    \item Constructing a hybrid Quantum Boltzmann Machine architecture: A hybrid model combines classical layers with a quantum layer that outputs latent features, forming the Quantum Boltzmann Machine (QBM).
    \item Analysing feature importance using gradient-based saliency maps: Gradients of the model output for the input are calculated from the quantum layer to reveal the most influential features.
\end{itemize}

\begin{figure}[htbp]
\centerline{\fbox{\includegraphics[scale=0.45]{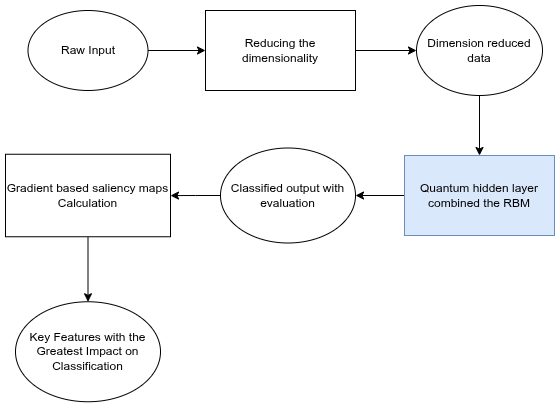}}}
\caption{Approach to integrate Quantum Computing with RBM}
\label{fig5}
\end{figure}

``Fig.~\ref{fig5}'' illustrates the steps in our approach to integrate Quantum Computing with RBM, which could improve the explainability.

\subsection{Reducing Dimensionality}
The research involved the binary classification task on the MNIST dataset, by restricting attention to two digits: 0 and 1. The choice is based on fairly different visual characteristics and common use for simplified benchmarking. Each image of 28×28 pixels was normalised, with values being the grey levels between 0 and 1. PCA \cite{abdi2010principal} was then applied to these 784-dimensional images, extracting four principal components to prepare the data for quantum computation and to reduce the input dimension.
\begin{figure}[htbp]
    \centerline{\fbox{\includegraphics[scale=0.30]{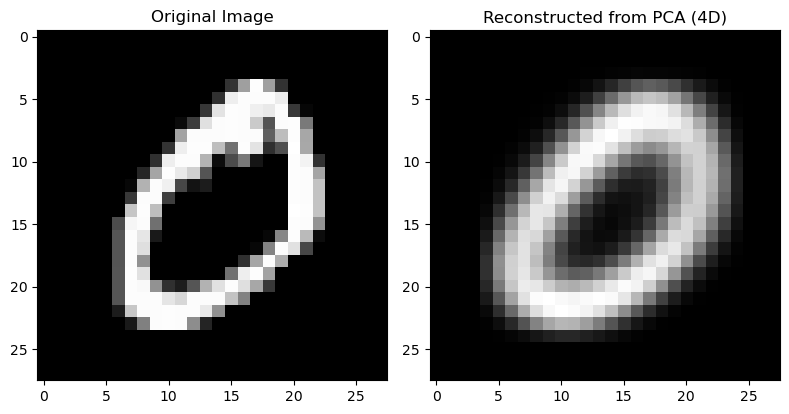}}}
    \caption{Before and after dimensionality reduction of ``0'' data from MNIST data set}
    \label{fig6}
\end{figure}

\subsection{Quantum Hidden Layer Design}
\begin{figure}[htbp]
    \centerline{\fbox{\includegraphics[scale=0.20]{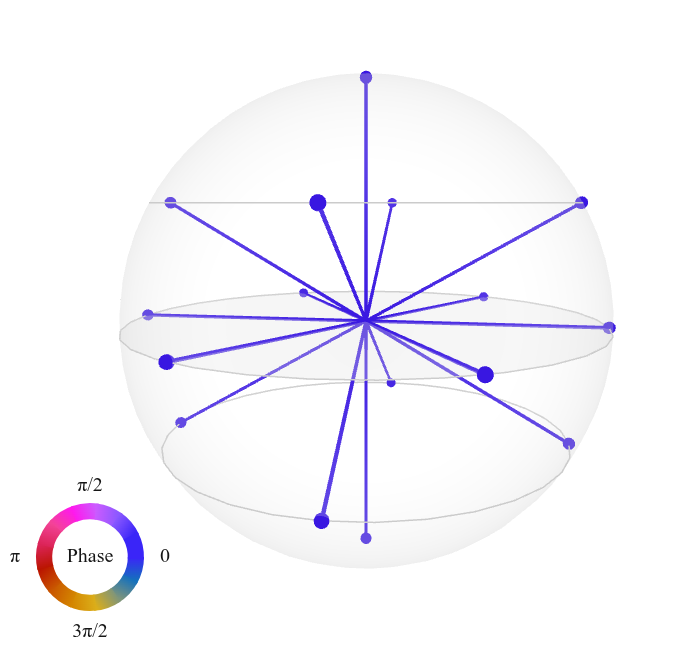}}}
    \caption{Q sphere representation of our quantum circuit using IBM quantum composer \cite{qiskit2024}}
    \label{fig7}
\end{figure}

A hybrid quantum-classical neural network layer is implemented using PennyLane to create a quantum-enhanced feature representation of classical input data. This is achieved through Angle Embedding \cite{lloyd2020quantum}, where each of the 4 input features is encoded as a rotation around the Y-axis (RY gate) on a corresponding qubit within the quantum circuit. These parameterised rotation gates allow for the classical input vector to be mapped into a valid quantum state using unitary transformations that abide by quantum mechanical rules. Through quantum encoding, this downstream quantum circuit can tackle the big entangled feature spaces through entanglement and parametrised gates, which can model subtle correlations that classical models simply cannot. Quantum measurements then return features which are fed into classical neural layers, keeping them together as a hybrid architecture..``Fig.~\ref{fig7}'' illustrates the q-sphere represents the state of a system of one or more qubits by associating each computational basis state with a point on the surface of a sphere \cite{qiskit2024}.

\begin{figure}[htbp]
    \centerline{\fbox{\includegraphics[scale=0.45]{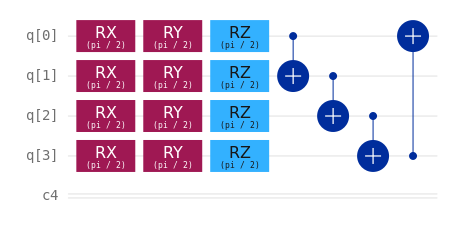}}}
    \caption{Quantum circuit representation of hidden layer \cite{qiskit2024}}
    \label{fig8}
\end{figure}

Following the embedding, we apply a strongly entangling layers template. This consists of multiple trainable layers, each composed of rotational gates applied to each qubit, followed by a series of entangling CNOT gates across adjacent qubits. These parameterised gates, denoted collectively as $V(\boldsymbol{\theta})$, allow the circuit to learn complex correlations by leveraging the quantum principles of superposition and entanglement. ``Fig.~\ref{fig8}'' shows the overview of the quantum circuit used in our hybrid-boltzmann machine.

Mathematically, the quantum state evolves as:
\begin{equation}
    |\psi(\mathbf{x}, \boldsymbol{\theta})\rangle = V(\boldsymbol{\theta}) \cdot U_{\text{embed}}(\mathbf{x}) \cdot |0\rangle^{\otimes 4}
\end{equation}

where $U_{\text{embed}}(\mathbf{x})$ represents the embedding unitary that encodes classical input $\mathbf{x}$ into quantum rotations or typically via angle embedding, and $V(\boldsymbol{\theta})$ is the parametrized quantum circuit\cite{du2020expressive} composed of entangling gates and rotation layers, controlled by trainable parameters $\boldsymbol{\theta}$. This circuit transforms the input into a rich quantum feature space, capturing nonlinear and entangled patterns in the data.

We extract quantum observables \cite{beltrametti1995quantum} by measuring the expectation values of the Pauli-Z operator on each qubit of the final quantum state:
\begin{equation}
    \mathbf{z} = \left[ \langle \psi | Z_0 | \psi \rangle, \langle \psi | Z_1 | \psi \rangle, \langle \psi | Z_2 | \psi \rangle, \langle \psi | Z_3 | \psi \rangle \right]
\end{equation}

This expectation value is another representation of the quantum-encoded input data \cite{rath2024quantum}. Now, having the data represented in this way, a classical neural network layer performs the final classification. The quantum feature extraction combined with classical processing forms the hybrid architecture of the quantum Boltzmann machine and hence provides more expressive representations, especially with very few data.

\subsection{Design of Classical Boltzmann Machine}
Hence, in our classical baseline, we design an RBM to learn hidden representations of the low-dimensional input features. The visible layer of the RBM comprises four nodes, each of which corresponds to one of the PCA-reduced MNIST input features. The hidden layer, on the other hand, has two binary stochastic units. The layers are interconnected symmetrically with no intra-layer connections, favouring accelerated learning through contrastive divergence.

The RBM defines a joint probability distribution\cite{osorio2024can} over visible units $\mathbf{v} \in \{0,1\}^4$ and hidden units $\mathbf{h} \in \{0,1\}^2$, parameterized by a weight matrix $\mathbf{W}$, and bias terms $\mathbf{b}$ (visible) and $\mathbf{c}$ (hidden):

\begin{equation}
    E(\mathbf{v}, \mathbf{h}) = -\mathbf{v}^\top \mathbf{W} \mathbf{h} - \mathbf{b}^\top \mathbf{v} - \mathbf{c}^\top \mathbf{h}
\end{equation}

This energy function defines a Boltzmann distribution\cite{yeturu2020machine}:

\begin{equation}
    P(\mathbf{v}, \mathbf{h}) = \frac{1}{Z} e^{-E(\mathbf{v}, \mathbf{h})}
\end{equation}

where $Z$ is the partition function, ensuring proper normalisation. Training is performed using contrastive divergence\cite{liu2013contrastive}, where the model approximates the gradient of the log-likelihood by comparing the data-dependent and model-dependent expectations through Gibbs sampling.

\begin{figure}[htbp]
    \centerline{\fbox{\includegraphics[scale=0.30]{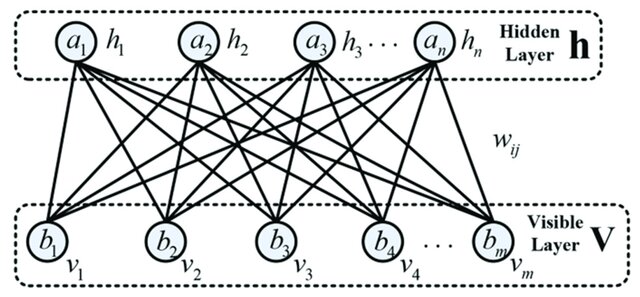}}}
    \caption{Visual representation of a restricted boltzmann machine \cite{fischer2012introduction}}
    \label{fig10}
\end{figure}

The CBM captures hidden input patterns in the binary input space by iteratively reconstructing inputs and correspondingly modifying weights to minimise reconstruction error. The learned hidden layer activations become compressed and abstract representations of the input, available for further analysis or classification. Namely, classical in the methods, the CBM forms a reasonable benchmark to assess which quantum-enhanced models can advance the field.

\subsection{Evaluation of QBM}
A binary classification subset of the MNIST dataset\cite{deng2012mnist}, considering only digits 0 and 1 to keep things simple for the quantum circuit input dimensionality, was utilised for the performance evaluation of the Quantum Boltzmann Machine (QBM). The original MNIST dataset contained 70,000 greyscale images of handwritten digits from 0 to 9. Out of those, only images labelled 0 or 1 were filtered. Each image consisted of 28×28 pixels, after flattening and then normalising into pixel intensity values between 0 and 1. To further reduce computational overheads and make it compatible with near-term quantum hardware constraints, PCA was used to squish the inputs down to 4 principal components only.

\begin{figure}[htbp]
    \centerline{\fbox{\includegraphics[scale=0.25]{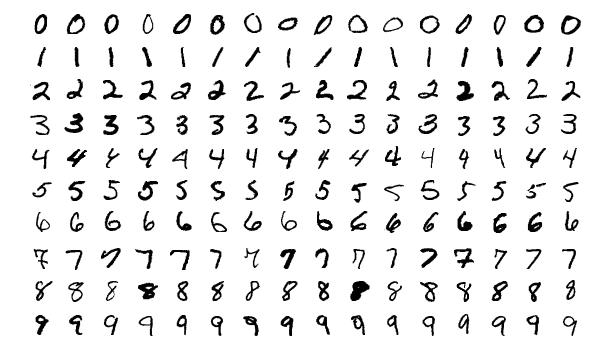}}}
    \caption{The MINIST dataset\cite{deng2012mnist}}
    \label{fig11}
\end{figure}

Both QBM and CBM were trained on PCA-reduced binary MNIST data with 50 epochs and a 0.01 learning rate. CBM was trained using Contrastive Divergence, and QBM using a hybrid quantum-classical training process. Both classification accuracy and model interpretability were emphasised in the test. For explainability, the feature relevance of the QBM was analysed with gradient-based saliency analysis, while the input contributions of the CBM were analysed with SHAP values. A t-SNE visualisation\cite{JMLR:v9:vandermaaten08a} of the hidden states of the QBM was also used to analyse how well the model learned class-specific representations. To quantify how each model distributed importance to input features, entropy was computed from the normalised saliency and SHAP values, revealing the sharpness or spread of the attention of each model across features. This multidimensional evaluation method attempted to compare not just prediction accuracy, but also quantum and classical model interpretability properties.

\section{Results and Discussion}
The Classical Boltzmann Machine and Quantum Boltzmann Machine had very different classification accuracy performance on binary-class MNIST data. The QBM operated at \textbf{83.5\%} accuracy, which was much better than \textbf{54.0\%} accuracy of the CBM. The significant boost suggests that the QBM is more capable of learning complex patterns in the reduced feature space, likely an effect of quantum-enhanced representational power.

\begin{figure}[htbp]
    \centerline{\fbox{\includegraphics[scale=0.32]{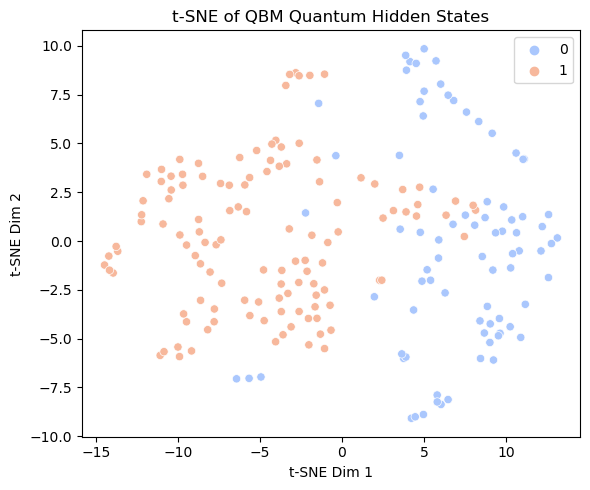}}}
    \caption{t-SNE of quantum hidden states}
    \label{fig12}
\end{figure}

To further investigate the acquired internal representations of the QBM, we have created a t-distributed Stochastic Neighbour Embedding (t-SNE)\cite{cieslak2020t} plot of the quantum hidden states. As shown in ``Fig.~\ref{fig12}'', the 2D visual representation strongly displays a two-digit category separation (0 and 1) with minimal overlap between clusters. This means that the QBM effectively inserts the input data into a latent space where the class-discriminative features become increasingly separable. The stable clustering behaviour confirms the model's capacity to learn informative and structured quantum embeddings, displaying high classification performance.

\begin{figure}[htbp]
    \centerline{\fbox{\includegraphics[scale=0.33]{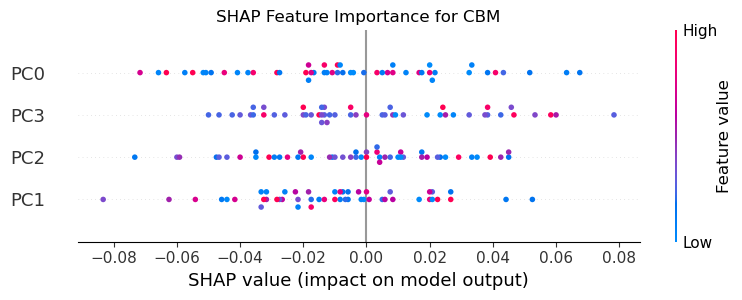}}}
    \caption{SHAP feature Attribution of CBM}
    \label{fig13}
\end{figure}

To explore the interpretability of the CBM, SHAP values were employed to quantify feature importance across the PCA-reduced input space. As indicated in ``Fig.~\ref{fig13}'', each principal component has various degrees of impact on the model output. Notably, the SHAP value distribution indicates that no single feature overwhelmingly controls the model's decision-making process. The colour gradient of the SHAP summary plot also depicts how different ranges of feature values affect positive or negative changes in the model's output, which yields fine-grained information regarding how the CBM operates on input data in binary classification.

\begin{figure}[htbp]
    \centerline{\fbox{\includegraphics[scale=0.3]{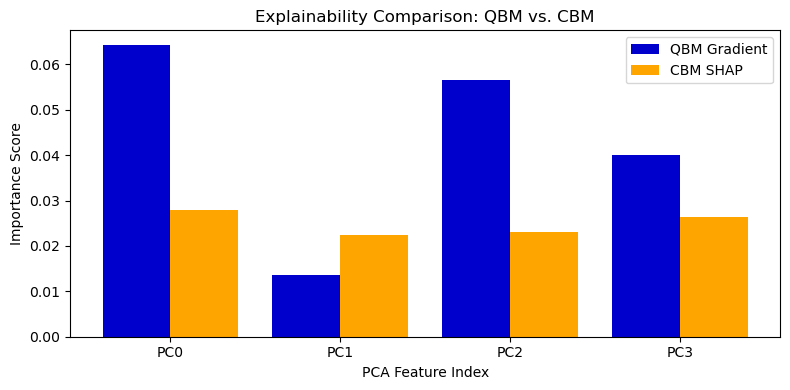}}}
    \caption{Comparison plot of QBM gradient vs CBM SHAP}
    \label{fig14}
\end{figure}

In ``Fig.~\ref{fig14}'', we present side-by-side feature importance attributions for the QBM via gradient-based saliency and the CBM via SHAP. Here, we see that the QBM places significantly more importance on \textbf{PC0} and \textbf{PC2}, as evidenced by their larger gradient magnitudes, while the CBM distributes importance more evenly across all four components. This distinction in feature importance points to fundamental differences in the internal representation and utilisation of information in quantum and classical models. Specifically, the more accurate attribution profiles of the QBM are reflective of a more focused internal representation, which can be attributed to its having higher classification accuracy. From an explainable AI perspective, these dominant features—PC0 and PC2 for the QBM here—can be considered \textbf{active ingredients} in the model's decision-making.

\begin{figure}[htbp]
    \centerline{\fbox{\includegraphics[scale=0.35]{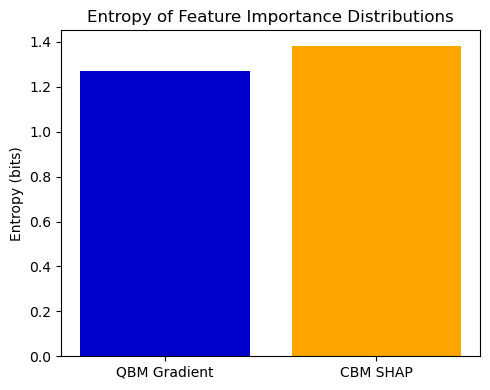}}}
    \caption{Entropy values of QBM and CBM}
    \label{fig15}
\end{figure}

To better understand how each model is focusing on the input features, we calculated the entropy of their feature importance values. In ``Fig.~\ref{fig15}'', we can see QBM had a lower entropy of \textbf{1.2704}, but the CBM had a higher entropy of \textbf{1.3820}. This shows that the QBM was focusing its attention on fewer but more relevant features, while the CBM was spreading its attention more densely over all features. The lower entropy in the QBM indicates that it performs better in isolating essential features—what we refer to as active ingredients—that underpin its decisions.

Overall, these results demonstrate that the QBM not only outperforms the CBM in classification accuracy but also offers clearer and more concentrated insights into which features drive its predictions. This strengthens its role in identifying the active ingredients within AI decision-making processes.

\section{Conclusion}
This research examined the use of Classical Boltzmann Machines (CBMs) and Quantum Boltzmann Machines (QBMs) for explainable AI on a binary MNIST classification task with PCA-reduced input. Central to our research was the comparison of their respective ``active ingredients'' for interpretability: SHAP values for CBMs and gradient-based saliency for QBMs. Whereas SHAP had provided traditional additive explanations in terms of cooperative game theory, the gradient-based method in QBMs revealed more holistic, globally important patterns of saliency governed by quantum properties such as entanglement and superposition.

Though the two models exhibited comparable classification performance, the QBM surpassed the SHAP concerning explainability richness. Saliency maps derived from the QBM offered richer explanations for the quantum energy landscape driving the model's behaviour, with the capacity to capture subtle feature interactions that the SHAP either omitted or oversimplified at times. This simply represents the core strength of QBMs: not only can they match classical models in predictiveness, but their quantum character allows for an intrinsically different, and perhaps more capable, path to model interpretability.

Therefore, the QBM stands as the better model in this comparative analysis, not just due to its classification performance, but through the interpretability afforded through its quantum-based active component. As machine learning based on quantum technology continues to evolve, the integration of embedded quantum explainability tools like gradient-based saliency will play a growing role in creating transparent and trustworthy AI systems. Future work will apply this approach to increasingly complex data sets, improve quantum circuit topologies, and potentially merge classical and quantum interpretability techniques into hybrid XAI systems.

\bibliographystyle{ieeetr}
\bibliography{references}

@incollection{el2015machine,
  title={What is machine learning?},
  author={El Naqa, Issam and Murphy, Martin J},
  booktitle={Machine learning in radiation oncology: theory and applications},
  pages={3--11},
  year={2015},
  publisher={Springer}
}

@article{hinton2007boltzmann,
  title={Boltzmann machine},
  author={Hinton, Geoffrey E},
  journal={Scholarpedia},
  volume={2},
  number={5},
  pages={1668},
  year={2007}
}

@article{patel2020overview,
  title={An overview of Boltzmann Machine and its special class},
  author={Patel, Anjali and Rama, Ranjith Kumar},
  journal={en. In},
  year={2020}
}

@inproceedings{alagiyawanna2024enhancing,
  title={Enhancing Small Dataset Classification Using Projected Quantum Kernels with Convolutional Neural Networks},
  author={Alagiyawanna, AMASD and Karunananda, Asoka and Mahasinghe, A and Silva, Thushari},
  booktitle={2024 8th SLAAI International Conference on Artificial Intelligence (SLAAI-ICAI)},
  pages={1--6},
  year={2024},
  organization={IEEE}
}

@article{amin2018quantum,
  title={Quantum boltzmann machine},
  author={Amin, Mohammad H and Andriyash, Evgeny and Rolfe, Jason and Kulchytskyy, Bohdan and Melko, Roger},
  journal={Physical Review X},
  volume={8},
  number={2},
  pages={021050},
  year={2018},
  publisher={APS}
}

@article{bergholm2018pennylane,
  title={Pennylane: Automatic differentiation of hybrid quantum-classical computations},
  author={Bergholm, Ville and Izaac, Josh and Schuld, Maria and Gogolin, Christian and Ahmed, Shahnawaz and Ajith, Vishnu and Alam, M Sohaib and Alonso-Linaje, Guillermo and AkashNarayanan, B and Asadi, Ali and others},
  journal={arXiv preprint arXiv:1811.04968},
  year={2018}
}

@inproceedings{jazaeri2019review,
  title={A review on quantum computing: From qubits to front-end electronics and cryogenic MOSFET physics},
  author={Jazaeri, Farzan and Beckers, Arnout and Tajalli, Armin and Sallese, Jean-Michel},
  booktitle={2019 MIXDES-26th International Conference" Mixed Design of Integrated Circuits and Systems"},
  pages={15--25},
  year={2019},
  organization={IEEE}
}

@incollection{NIPS2017_7062,
title = {A Unified Approach to Interpreting Model Predictions},
author = {Lundberg, Scott M and Lee, Su-In},
booktitle = {Advances in Neural Information Processing Systems 30},
editor = {I. Guyon and U. V. Luxburg and S. Bengio and H. Wallach and R. Fergus and S. Vishwanathan and R. Garnett},
pages = {4765--4774},
year = {2017},
publisher = {Curran Associates, Inc.},
url = {http://papers.nips.cc/paper/7062-a-unified-approach-to-interpreting-model-predictions.pdf}
}

@article{mackiewicz1993principal,
  title={Principal components analysis (PCA)},
  author={Ma{\'c}kiewicz, Andrzej and Ratajczak, Waldemar},
  journal={Computers \& Geosciences},
  volume={19},
  number={3},
  pages={303--342},
  year={1993},
  publisher={Elsevier}
}

@article{abdi2010principal,
  title={Principal component analysis},
  author={Abdi, Herv{\'e} and Williams, Lynne J},
  journal={Wiley interdisciplinary reviews: computational statistics},
  volume={2},
  number={4},
  pages={433--459},
  year={2010},
  publisher={Wiley Online Library}
}

@article{lloyd2020quantum,
  title={Quantum embeddings for machine learning},
  author={Lloyd, Seth and Schuld, Maria and Ijaz, Aroosa and Izaac, Josh and Killoran, Nathan},
  journal={arXiv preprint arXiv:2001.03622},
  year={2020}
}

@misc{qiskit2024,
      title={Quantum computing with {Q}iskit},
      author={Javadi-Abhari, Ali and Treinish, Matthew and Krsulich, Kevin and Wood, Christopher J. and Lishman, Jake and Gacon, Julien and Martiel, Simon and Nation, Paul D. and Bishop, Lev S. and Cross, Andrew W. and Johnson, Blake R. and Gambetta, Jay M.},
      year={2024},
      doi={10.48550/arXiv.2405.08810},
      eprint={2405.08810},
      archivePrefix={arXiv},
      primaryClass={quant-ph}
}

@article{beltrametti1995quantum,
  title={Quantum observables in classical frameworks},
  author={Beltrametti, EG and Bugajski, S{\l}awomir},
  journal={International Journal of Theoretical Physics},
  volume={34},
  pages={1221--1229},
  year={1995},
  publisher={Springer}
}

@article{du2020expressive,
  title={Expressive power of parametrized quantum circuits},
  author={Du, Yuxuan and Hsieh, Min-Hsiu and Liu, Tongliang and Tao, Dacheng},
  journal={Physical Review Research},
  volume={2},
  number={3},
  pages={033125},
  year={2020},
  publisher={APS}
}

@article{rath2024quantum,
  title={Quantum data encoding: A comparative analysis of classical-to-quantum mapping techniques and their impact on machine learning accuracy},
  author={Rath, Minati and Date, Hema},
  journal={EPJ Quantum Technology},
  volume={11},
  number={1},
  pages={72},
  year={2024},
  publisher={Springer Berlin Heidelberg}
}

@article{osorio2024can,
  title={Can a Hebbian-like learning rule be avoiding the curse of dimensionality in sparse distributed data?},
  author={Os{\'o}rio, Maria and Sa-Couto, Luis and Wichert, Andreas},
  journal={Biological Cybernetics},
  volume={118},
  number={5},
  pages={267--276},
  year={2024},
  publisher={Springer}
}

@incollection{yeturu2020machine,
  title={Machine learning algorithms, applications, and practices in data science},
  author={Yeturu, Kalidas},
  booktitle={Handbook of statistics},
  volume={43},
  pages={81--206},
  year={2020},
  publisher={Elsevier}
}

@inproceedings{liu2013contrastive,
  title={Contrastive divergence learning for the restricted Boltzmann machine},
  author={Liu, Jian-Wei and Chi, Guang-Hui and Luo, Xiong-Lin},
  booktitle={2013 Ninth international conference on natural computation (ICNC)},
  pages={18--22},
  year={2013},
  organization={IEEE}
}

@inproceedings{fischer2012introduction,
  title={An introduction to restricted Boltzmann machines},
  author={Fischer, Asja and Igel, Christian},
  booktitle={Iberoamerican congress on pattern recognition},
  pages={14--36},
  year={2012},
  organization={Springer}
}

@article{deng2012mnist,
  title={The mnist database of handwritten digit images for machine learning research},
  author={Deng, Li},
  journal={IEEE Signal Processing Magazine},
  volume={29},
  number={6},
  pages={141--142},
  year={2012},
  publisher={IEEE}
}

@article{JMLR:v9:vandermaaten08a,
  author  = {Laurens van der Maaten and Geoffrey Hinton},
  title   = {Visualizing Data using t-SNE},
  journal = {Journal of Machine Learning Research},
  year    = {2008},
  volume  = {9},
  number  = {86},
  pages   = {2579--2605},
  url     = {http://jmlr.org/papers/v9/vandermaaten08a.html}
}

@article{cieslak2020t,
  title={t-Distributed Stochastic Neighbor Embedding (t-SNE): A tool for eco-physiological transcriptomic analysis},
  author={Cieslak, Matthew C and Castelfranco, Ann M and Roncalli, Vittoria and Lenz, Petra H and Hartline, Daniel K},
  journal={Marine genomics},
  volume={51},
  pages={100723},
  year={2020},
  publisher={Elsevier}
}

\end{document}